\documentclass[review]{elsarticle}

\usepackage{color}
\usepackage{amsmath,amsfonts,amssymb,pifont}
\usepackage{booktabs}
\usepackage{graphicx}
\usepackage{bm}
\usepackage{array}
\usepackage{subfigure}
\usepackage{multirow}
\usepackage{appendix}
\usepackage[colorlinks=true, allcolors=blue]{hyperref}
\usepackage[table,xcdraw]{xcolor}
\usepackage{lineno,hyperref}
\usepackage{wrapfig}

\journal{Journal of \LaTeX\ Templates}

\begin{document}

\begin{frontmatter}

\title{ICAFusion: Iterative Cross-Attention Guided Feature Fusion for Multispectral Object Detection}

\author[a]{Jifeng~Shen\corref{mycorrespondingauthor}}
\cortext[mycorrespondingauthor]{Corresponding author}
\ead{shenjifeng@ujs.edu.cn}

\author[a]{Yifei~Chen}
\author[a]{Yue~Liu}
\author[b]{Xin~Zuo}
\author[c]{Heng~Fan}
\author[d]{Wankou~Yang}
\address[a]{School of Electrical and Information Engineering, Jiangsu University, Zhenjiang, 212013, China}
\address[b]{School of Computer Science and Engineering, Jiangsu University of Science and Technology, Zhenjiang, 212003, China}
\address[c]{Department of Computer Science and Engineering, University of North Texas, Denton, TX 76207, USA}
\address[d]{School of Automation, Southeast University, Nanjing, 210096, China}


\begin{abstract}
Effective feature fusion of multispectral images plays a crucial role in multispectral object detection. Previous studies have demonstrated the effectiveness of feature fusion using convolutional neural networks, but these methods are sensitive to image misalignment due to the inherent deficiency in local-range feature interaction resulting in the performance degradation. To address this issue, a novel feature fusion framework of dual cross-attention transformers is proposed to model global feature interaction and capture complementary information across modalities simultaneously. This framework enhances the discriminability of object features through the query-guided cross-attention mechanism, leading to improved performance. However, stacking multiple transformer blocks for feature enhancement incurs a large number of parameters and high spatial complexity. To handle this, inspired by the human process of reviewing knowledge, an iterative interaction mechanism is proposed to share parameters among block-wise multimodal transformers, reducing model complexity and computation cost. The proposed method is general and effective to be integrated into different detection frameworks and used with different backbones. Experimental results on KAIST, FLIR, and VEDAI datasets show that the proposed method achieves superior performance and faster inference, making it suitable for various practical scenarios. Code will be available at https://github.com/chanchanchan97/ICAFusion.
\end{abstract}


\end{frontmatter}


\section{INTRODUCTION}
\label{sec:intro}

Thermal spectrum range image provides a special way to perceive the natural scenes, which is believed to complement visible spectrum images in computer vision. Multispectral image feature representation and fusion is a challenging problem, serving a variety of downstream vision tasks, such as object detection, semantic segmentation and object tracking. As a fundamental vision task, object detection remains a hot topic in both academia and industry, and has made considerable progress in  small object detection \cite{sod2023pr}, face and pedestrian detection \cite{csp2023pr} and oriented object detection \cite{cheng2018learning, xie2021oriented}, thanks to the rapid development of convolutional neural networks. However, these methods are still vulnerable to environmental factors, such as severe weather conditions and changing illumination. In order to improve the robustness and accuracy of object detectors in all-weather conditions, multispectral object detection based on both RGB and thermal images has become a viable solution and is gaining popularity in recent academic studies.

\begin{figure*}[!t]
	\centering
	\setlength{\abovecaptionskip}{0.cm}
	\subfigure[]{
		\begin{minipage}[b]{0.4\linewidth}
			\includegraphics[width=1\linewidth]{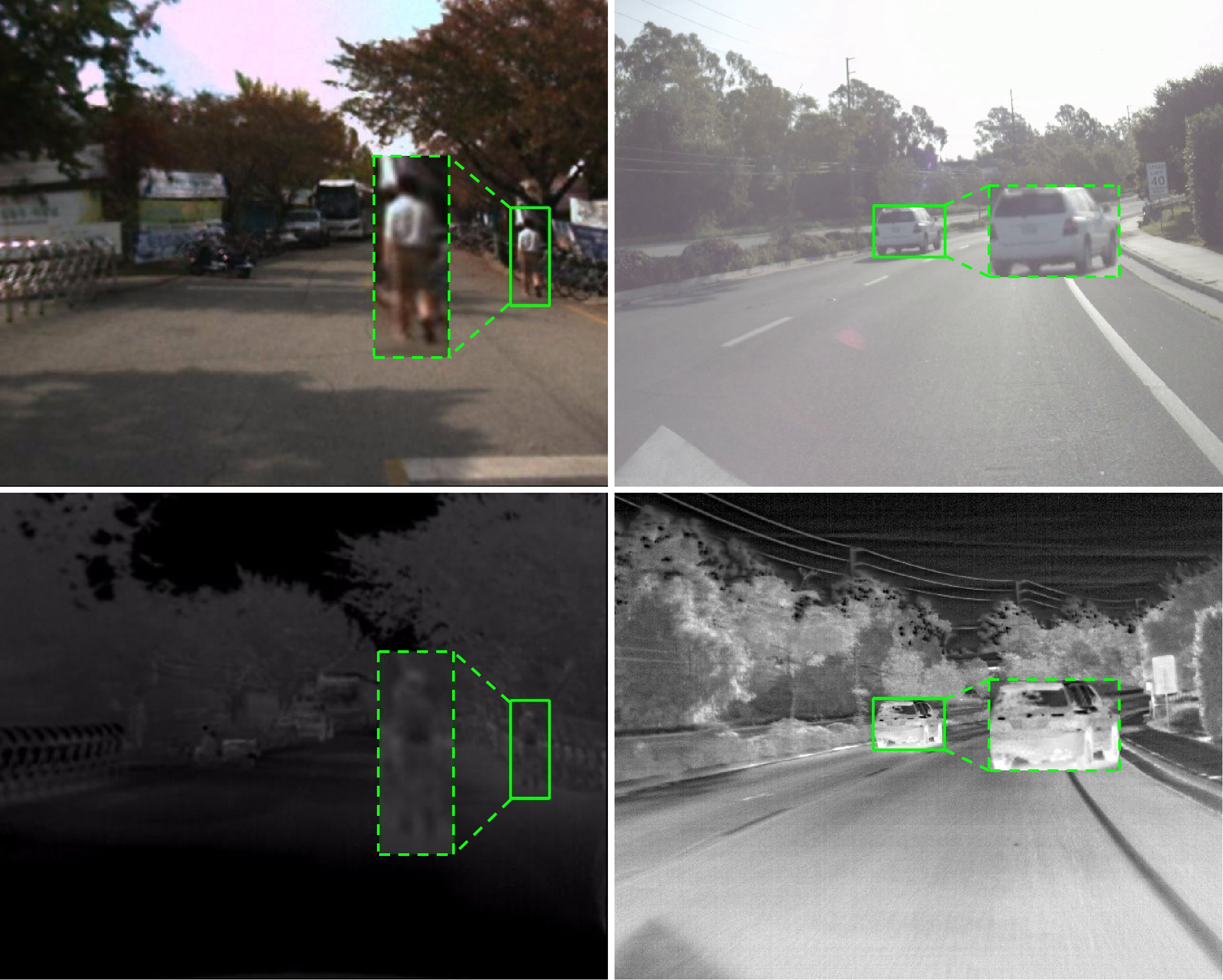}
		\end{minipage}
	}\hspace{-3mm}
	\subfigure[]{
		\begin{minipage}[b]{0.4\linewidth}
			\includegraphics[width=1\linewidth]{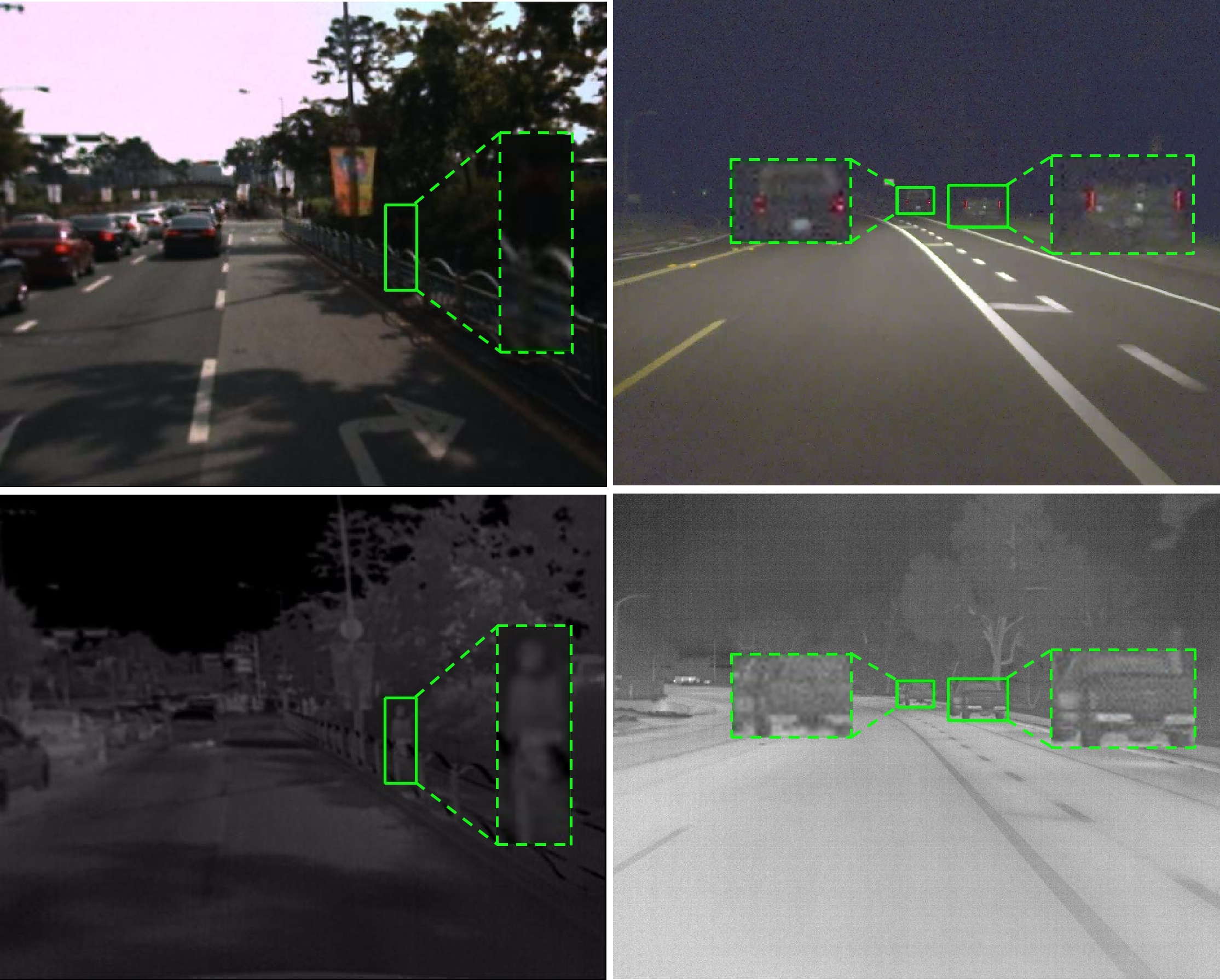}
		\end{minipage}
	}\hspace{-3mm}
	\caption{Visualization of objects in paired RGB-thermal images. The first and second rows are RGB and thermal images respectively. The objects are more perceptible in RGB images (a), while they are much easier to observe in thermal images (b). The dashed windows are cropped and enlarged for better visualization.}
	\label{fig:Fig1}
\end{figure*}

Compared with mono-modal object detection, the use of multiple modalities in object detection can provide a richer visual representation of objects, enabling them to effectively compensate for each other. As illustrated in Fig.~\ref{fig:Fig1}(a), RGB images can provide detailed colors, textures, and contours of objects under good illumination conditions, but these features may not be visible in thermal images. In contrast, Fig.~\ref{fig:Fig1}(b) shows that in poor illumination conditions, such as night or dark places. It is difficult to distinguish object details and edges from the background in RGB images, but thermal images can still provide perceptible contour features due to their unique energy radiation imaging mechanism. These characteristics indicate the complementary nature of RGB and thermal modalities. As a result, effective feature fusion from different modalities is critical for multispectral object detection.

In the previous studies, convolution-based feature fusion has been widely used in current state-of-the-art methods. The pioneering work \cite{liu2016multispectral} has explored different convolutional neural network (CNN) based fusion architectures and has shown that halfway fusion can lead to desirable performance. Zhang et al. \cite{zhang2019cross} encode the interaction between different modalities and fuse features adaptively. Zhang et al. \cite{zhang2021guided} introduce auxiliary pedestrian masks to guide the feature representation of inter-modal and intra-modal. Fu et al. \cite{fu2021adaptive} introduce the attention modules to perform the pixel-level feature fusion from the spatial and channel dimension. However, convolution-based feature fusion only pays attention to local feature information, and it lacks the ability to model long-range feature relationships due to the limited receptive field of CNN. To this end, we propose a novel dual cross-attention transformer fusion method that aggregates the feature information of RGB and thermal modalities from both local and global perspectives. Although transformers have the potential to address this limitation of CNNs, we observe that naive usage in feature fusion can lead to massive feature redundancy, resulting in excessive computational load and memory requirements. To address these issues, we have carefully studied the structure of fusion transformers and have attempted to answer two key questions.

\textbf{How to effectively borrow complementary information from other modalities ?}
The performance of multispectral object detection is strongly correlated with the quality of feature fusion. The traditional fusion method relied on feature concatenation or addition, which is susceptible to image misalignment due to the inherent constraint of limited local-range feature interaction. Inspired by pretraining cross-modal feature representations in vision-language tasks \cite{lu2019vilbert}, we have proposed a cross-attention fusion transformer for multispectral image feature fusion to tackle this issue. This method is designed to capture the complementary features from other modalities, and specifically tailored to enhance both feature branches simultaneously. Additionally, our proposed fusion transformer naturally benefits from its long-range modeling feature interactions, which aid in the discovery of discriminative complementing information from other modalities. 
Different from the single  transformer fusion methods \cite{qingyun2021cross} that concatenate the tokens of each modality and compute the queries, keys, values from all modality information, our proposed method computes the correlation across modalities only with queries from the auxiliary modality.

\textbf{How to efficiently integrate and refine multispectral image features~?}
Transformer-based models are well known in the vision world for their enormous computational complexity. Besides, most of the existing methods \cite{dosovitskiy2020image} stack numerous blocks to boost performance, resulting in a surge of computational cost. However, humans generally repeatedly review the knowledge after learning new ones, which aids in the retention of what they have learned. Inspired by this, we have proposed an iterative learning strategy. This method not only learns global complementary information based on the bidirectional feature flow interaction between the RGB and thermal branches, but also iteratively refines the feature representation of inter-modality and intra-modality simultaneously, thereby strengthening the discriminative feature information. In contrast to standard methods that stack multiple blocks, our proposed iterative learning strategy shares parameters in each block and improves the balance between model performance and complexity.

To summarize, our main contributions are as follows:
\begin{itemize}
	\item A novel dual cross-attention feature fusion method is proposed for multispectral object detection, which simultaneously aggregates complementary information from RGB and thermal images.
	\item An iterative learning strategy is tailored for efficient multispectral feature fusion, which further improves the model performance without additional increase of learnable parameters.
	\item The proposed feature fusion method is both generalizable and effective, which can be plugged into different backbones and equipped with different detection frameworks.
	\item The proposed CFE/ICFE module can function with different input image modalities, which provide a feasible solution when one of the modality is missing or has pool quality.
	
	\item The proposed method can achieve the state-of-the-arts results on KAIST \cite{hwang2015multispectral}, FLIR \cite{flir_dataset} and VEDAI \cite{razakarivony2016vehicle} datasets, while also obtains very fast inference speed.
\end{itemize}

The rest of this paper is organized as follows. Section II introduces the related work published in recent years. Section III describes our proposed method in detail. The experimental results are given in Section IV and we conclude the paper in Section V.

\section{Related work}
\subsection{Multispectral Object Detection}
Recent multispectral object detection research has made consistent progress, particularly in multispectral pedestrian detection. 
Hwang et al.\cite{hwang2015multispectral} have built the first multispectral pedestrian benchmark and provided a hand-crafted approach based on the Aggregated Channel Feature (ACF) by extending the infrared channel features. To make full use of complementary information between different modalities, Zhou et al. \cite{zhou2020improving} use the ethos of differential amplifiers to understand the consistency and difference between distinct modalities by leveraging common-mode and differential-mode information. Zhang et al. \cite{zhang2020multispectral} employ a method for cyclically fusing and refining multispectral features, which aims to improve the consistency of both modalities. To capture the discriminative object features in the multispectral images, MSDS-RCNN \cite{limultispectral} is the first work that uses semantic segmentation to guide multispectral object detection via multi-task learning. Shen et al. \cite{shen2022mask} propose a mask-guided mutual attention module and score fusion module based on the anchor-free detector, which achieves the trade-off between accuracy and speed. Taking into account the variation in lighting between day and night, Li et al. \cite{li2019illumination} have developed an Illumination-aware Network (IAN) that predicts an illumination weight from RGB images via a Gate Unit and weights the results from the RGB and thermal branches. With the goal of addressing the misalignment problem between two modalities, Zhang et al. \cite{zhang2019weakly} utilize a Region Feature Alignment (RFA) module to anticipate feature offset between RGB and thermal pictures. Kim et al. \cite{kim2021uncertainty} take into account the model's many uncertainties and provided a loss function to direct the visual representation of two modalities in the feature-level to be similar. To tackle the issue that the existing methods lack the ability to model long-range dependencies across modalities, CFT \cite{qingyun2021cross} and LGADet \cite{zuo2022lgadet} are proposed to improve the quality of feature fusion with local and global attention mechanisms. However, these methods only simply integrate Transformer or non-local network into detection framework without taking full advantage of complementary information between global features. In this paper, we propose a novel iterative cross-attention interaction method that fully utilizes both local and global feature information between different modalities.

\subsection{Attention-based methods}
Attention mechanism originates from research on human vision, which is widely used in the computer vision field. SENet \cite{hu2018squeeze} proposes a simple yet effective structure to learn the weights between different channels with fully connected network.  Inspired by this, SKNet \cite{li2019selective} proposes a dynamic selection mechanism that allows each neuron to adaptively adjust its receptive field size based on multiple scales of input information. CBAM \cite{woo2018cbam} proposes a lightweight and general module to adaptively refine the features in both channel and spatial dimensions. 
ECANet \cite{wang2020eca} proposes a local cross-channel interaction strategy with an adaptive one-dimensional convolution, which only involves a handful of parameters while bringing clear performance gain. 
More recently, CANet \cite{cheng2023class} is proposed with an effective class-specific attention encoding module that learns a class-specific dictionary to encode class attention maps. In this paper, we propose a cross-modal attention module which leverages the complementary information from auxiliary modality to enhance the mono-modality feature representation.

\subsection{Transformer for Multimodal Learning}
Transformer has been applied to multimodal tasks as a result of its significant performance improvement in NLP and CV.  Multi-Modality Cross Attention (MMCA) \cite{wei2020multi} is proposed for image and text matching, which jointly models the intra-modal and inter-modal relationships between image and sentence in a unified depth model. 
TransFusion \cite{bai2022transfusion} provides a robust solution for LiDAR-camera fusion with a soft-association mechanism to handle inferior image conditions. 
Botach et al. \cite{botach2022end} propose an architecture of multimodal tracking transformer, which models referring video object segmentation task as sequence prediction problems.
A token-based multi-task decoder \cite{liu2021visual} for RGB-D salient object detection method is developed by introducing task-related tokens and a novel patch-task-attention mechanism. 
Li et al. \cite{li2021trear} propose a Transformer-based RGB-D egocentric action recognition framework and  model the temporal structure of the data from different modalities with self-attention. 
Xiao et al. \cite{xiao2022attribute} have designed five attribute-specifc fusion branches to integrate RGB and thermal features under various challenges of RGB-T tracking, and strengthened the aggregated feature and modality-specifc features with an enhancement fusion transformer. 
These studies have proved that Transformer is effective in various multimodal tasks. 
In this paper, we introduce Transformer into multispectral object detection that aims to better harvest the complementary information between RGB and thermal modality from a global perspective.

\section{The proposed method}

\begin{figure}[]
	\centering
	\setlength{\abovecaptionskip}{0.cm}
	\includegraphics[width=0.8\linewidth]{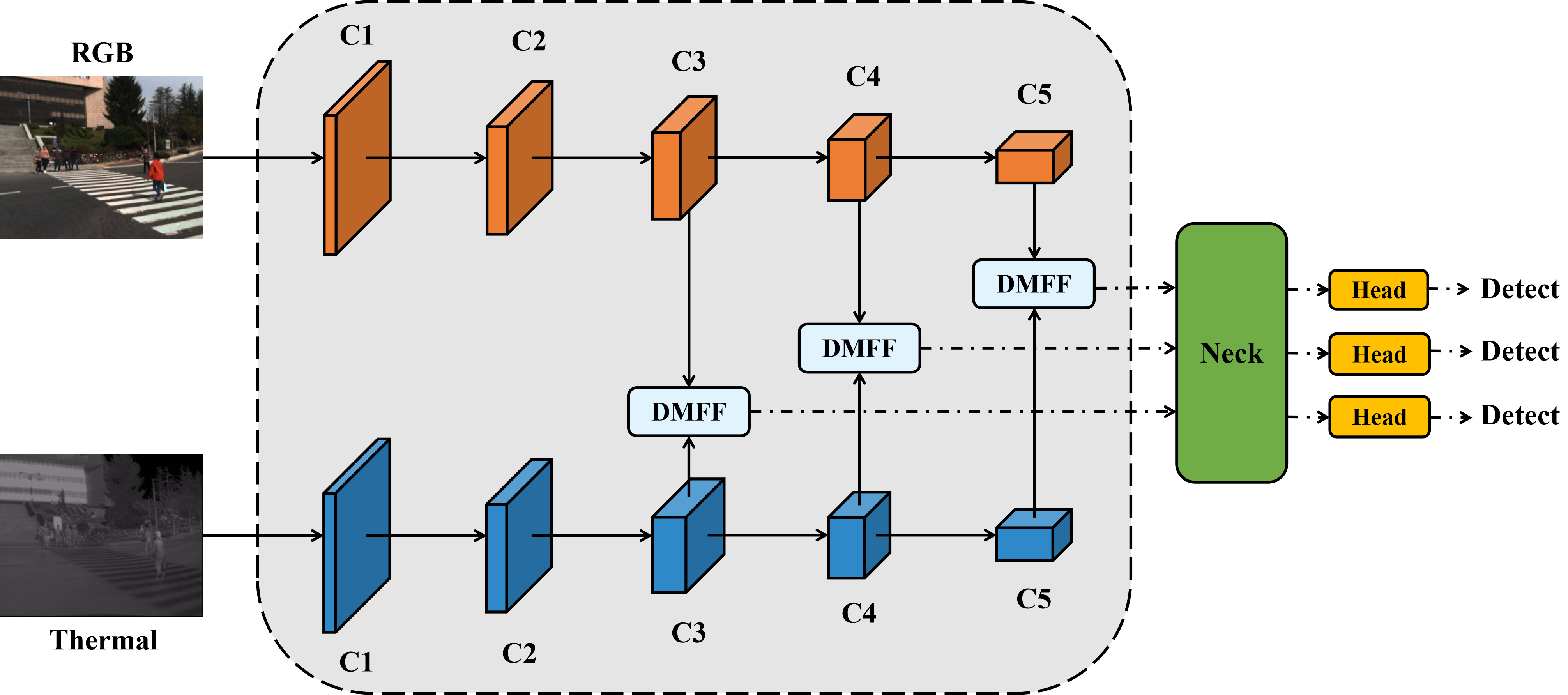}
	\caption{Overview of our multispectral object detection framework. (The upper and bottom branch are the RGB and thermal feature extraction module, C1$\sim$C5 represent different scales of feature maps, DMFF module is our proposed feature fusion method, Neck module is the multiscale feature aggregation network, and Head module outputs the final detection results.)}
	\label{fig:Fig2}
\end{figure}

\subsection{Architecture}
As illustrated in Fig.~\ref{fig:Fig2}, the proposed method is a dual-branch backbone network, which is tailored for feature extraction from RGB-thermal image pairs. Our method mainly comprises of three stages: the mono-modal feature extraction, the dual-modal feature fusion and the detection neck and head. The mono-modal feature extraction is firstly used for RGB and thermal images independently, which can be formulated in Eq.~\ref{eq1}.
\begin{equation}
\begin{aligned}
\bm{F}^{i}_{R} = \Psi_{backbone}(\bm{I}_R;\bm{\theta}_R), 
\bm{F}^{i}_{T} = \Psi_{backbone}(\bm{I}_T;\bm{\theta}_T)
\end{aligned}
\label{eq1}
\end{equation}

\noindent where $\bm{F}^{i}_{R}$, $\bm{F}^{i}_{T} \in \mathbb{R}^{W\times H\times C}$ denote the feature maps from the i-th layer (i=3, 4, 5) of the RGB and thermal branch respectively.  $H$, $W$ and $C$ denote height, width and channel number of feature maps. $\bm{I}_{R}$, $\bm{I}_{T} \in \mathbb{R}^{W\times H\times C}$ represent the input RGB and thermal images, $\Psi_{backbone}$ denotes the feature extraction function with parameters $\bm{\theta_R}$ and $\bm{\theta_T}$  for RGB and thermal branch respectively. 
In generic object detection, VGG16 \cite{simonyan2014very}, ResNet \cite{he2016deep} and CSPDarkNet \cite{bochkovskiy2020yolov4} are commonly used as function $\Psi_{backbone}$.
In the feature extraction stage, the multi-scale features are commonly utilized to capture objects with different size. 

Secondly, given feature maps of $\bm{F}^{i}_{R}$ and $\bm{F}^{i}_{T}$, cross-modal feature fusion is required to aggregate features from different branches in multispectral object detection, which can be defined in Eq.~\ref{eq3}.
\begin{equation}
\begin{aligned}
\bm{F}^{i}_{R+T} = \Phi_{fusion}(\bm{F}^i_R;\bm{F}^i_T; \bm{\theta_f})
\end{aligned}
\label{eq3}
\end{equation}
\noindent where $\bm{F}^{i}_{R+T} \in \mathbb{R}^{W\times H\times C}$ denotes the fused features in the i-th layer. $\Phi_{fusion}(\cdot)$ denotes the feature fusion function with parameter $\bm{\theta_f}$.
Given that previous researches \cite{liu2016multispectral,li2019illumination} have explored
different fusion architectures and validated that halfway fusion outperforms the other fusion methods, we use halfway fusion as the default setting and fuse the multimodal features from convolution layer C3$\sim$C5 as shown in Fig. \ref{fig:Fig2}.
In general, addition operation or NIN fusion \cite{limultispectral} \footnote{${{\Phi}_{fusion}=conv_{1\times1}([\bm{{F}}_{R},\bm{{F}}_{T}])}$, where $conv_{1\times1}$ is a $1\times1$ convolution, $[\cdot]$ denotes the concatenate operation.} is commonly used as feature fusion function $\Phi_{fusion}(\cdot)$. In this paper, a dual cross-attention feature fusion transformer is proposed to model $\Phi_{fusion}(\cdot)$, which will be described in section \ref{section:3.2}.

Finally, the feature maps from $\{\bm{F}^i_{R+T}\}^L_{i=1}$ are fed to the detector neck for multi-scale feature fusion, and then delivered to the detector head for subsequent classification and regression which is formulated in Eq.~\ref{eq4}.
\begin{equation}
\begin{aligned}
{[\bm{D}_{cls}, \bm{D}_{bbox}] = 
\phi_{head}(\phi_{neck}(\{\bm{F}^i_{R+T}\}^L_{i=1});\bm{\theta_h})}
\end{aligned}
\label{eq4}
\end{equation}

\noindent where $\phi_{neck}$ and $\phi_{head}$ represent the multi-scale feature aggregation and detection head function.  FPN \cite{lin2017feature} and PANet \cite{liu2018path} is commonly utilized as function $\phi_{neck}$ to enhance the semantic expression and localization ability of features, while $\phi_{head}$ acts as a role of classification and bounding box regression with parameter $\bm{\theta_h}$, such as detection head of YOLO \cite{bochkovskiy2020yolov4} and FCOS \cite{tian2019fcos}. 
For a fair comparison, we adopt these default setting of detection necks and heads in the original paper.

\begin{figure*}[!t]
	\centering
	\setlength{\abovecaptionskip}{0.cm}
	\includegraphics[width=.9\linewidth]{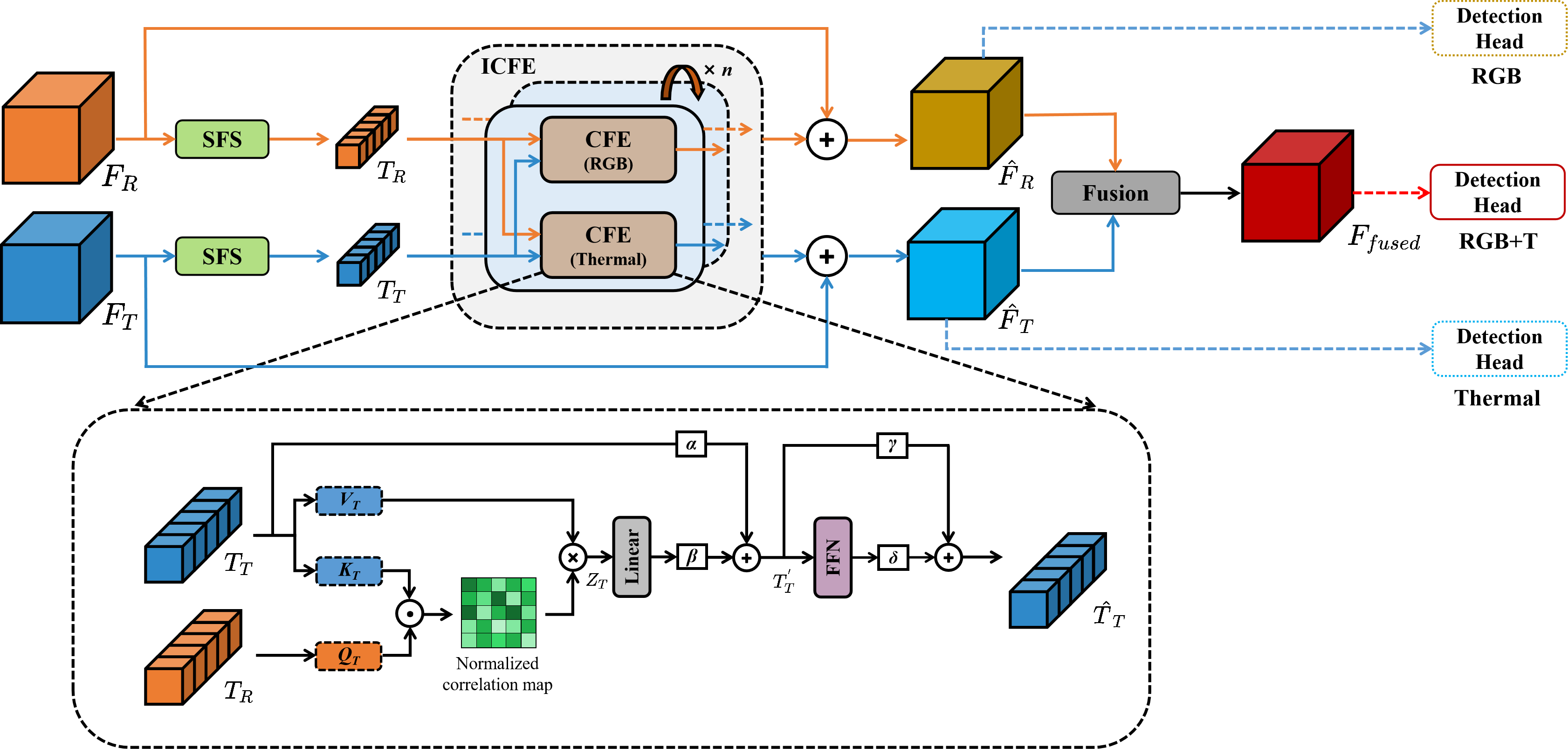}
	\caption{Illustration of the proposed DMFF module.(In the upper row, the proposed DMFF module comprised of Spatial Feature Shrinking (SFS) module, Iterative Cross-modal Feature Enhancement (ICFE) module and the bimodal feature fusion module with NIN fusion. The SFS module compresses the size of the feature map for subsequent CFE module, the ICFE module refines the cross-modal features by dual CFE modules, and the bimodal feature fusion module conducts the local feature fusion from the output of ICFE module. The bottom row illustrates the details of the CFE module for the enhancement of thermal modality.)}
	\label{fig:Fig3}
\end{figure*}

\subsection{Dual-modal Feature Fusion (DMFF)}
\label{section:3.2}
Fig.~\ref{fig:Fig3} illustrates the structure of our Dual-modal Feature Fusion (DMFF) module, which mainly contains three components: the Spatial Feature Shrinking (SFS) module, the Iterative Cross-modal Feature Enhancement (ICFE) module, and the bimodal feature fusion module with NIN fusion. The modules will be detailed in the following sections.

\subsubsection{Cross-modal Feature Enhancement (CFE)}
Different from the previous studies which capture the local features of different modalities, the proposed CFE module enables mono-modal to learn more complementary information from auxiliary modality in a global perspective. The proposed CFE module not only retrieves the complementary relationship between the RGB and thermal modality, but also overcome the deficiency in modeling the long-range dependencies of cross-modal features.

Given input feature maps $\bm{{F}}_{R}$ and $\bm{{F}}_{T} \in \mathbb{R}^{H\times W\times C}$\footnote{The vector $\bm{{F}}_{R}$ and $\bm{{F}}_{T}$ represent the feature maps from the i-th layer of different branches, similar to that in Eq. \ref{eq1}. For simplicity, we remove the superscript $i$.}, we first flatten each feature map into a set of tokens and add a learnable position embedding, which is a trainable parameter of dimension $HW \times C$ that encodes the spatial information between different tokens. After that, we can obtain a set of tokens $\bm{{T}}_{R}, \bm{{T}}_{T} \in \mathbb{R}^{HW\times C}$ with position embeddings as the input of CFE module. Since the RGB-thermal image pairs are usually not perfectly aligned, we employ dual CFE modules to harvest the complementary information for the enhancement of the RGB and thermal features respectively. The parameters are not shared between the two CFE modules. In Fig.~\ref{fig:Fig3}(bottom), we only illustrate the CFE module of thermal branch for clarity, which is formulated in Eq.~\ref{eq5}.
\begin{equation}
\begin{aligned}
\hat{\bm{T}}_{T}=\mathcal{F}_{CFE-T}(\{\bm{{T}}_{R},\bm{{T}}_{T}\})
\end{aligned}
\label{eq5}
\end{equation}

\noindent where $\bm{{T}}_{R}$ and $\bm{{T}}_{T}$ denotes the RGB and thermal feature tokens input to the CFE module. $\hat{\bm{T}}_{T}$ indicates the enhanced thermal features with the CFE module. $\mathcal{F}_{CFE-T}(\cdot)$ denotes our proposed CFE module for thermal branches. 

The details of the CFE module are as follows. Firstly, the tokens of thermal modality $\bm{{T}}_{T}$ are projected to two separate matrices $\bm{{V}}_T, \bm{{K}}_T \in \mathbb{R}^{HW \times C}$ to compute a set of values and keys (Eq. \ref{eq6}). And then, the tokens of RGB modality $\bm{{T}}_{R}$ are projected to another separate matrix $\bm{Q}_{R} \in \mathbb{R}^{HW \times C}$ to compute a set of queries (Eq. \ref{eq6}).
\begin{equation}
\begin{aligned}
\bm{{V}}_{T}=\bm{{T}}_{T}\bm{W}^V, 
\bm{{K}}_{T}=\bm{{T}}_{T}\bm{W}^K,
\bm{Q}_{R}=\bm{{T}}_{R}\bm{W}^Q
\end{aligned}
\label{eq6}
\end{equation}

\noindent where $\bm{W}^V$, $\bm{W}^K$  and $\bm{W}^Q \in \mathbb{R}^{C \times C}$ denote the weight matrices.

Secondly, the correlation matrix is built via dot-product operation, followed by a softmax function normalizes the correlation scores, which represents the similarity between different features of RGB and thermal modality. After that, the vector $\bm{Z}_T$ is obtained by multiplying the correlation matrix with vector $\bm{{V}}_T$ (Eq.~\ref{eq9}), which refines the RGB features by leveraging the similarity across modalities. 
\begin{equation}
\begin{aligned}
\bm{Z}_T={\rm softmax}(\frac{\bm{Q}_R{\bm{{K}}_T}^T}{\sqrt{D_K}}) \cdot \bm{{V}}_T
\end{aligned}
\label{eq9}
\end{equation}
\begin{equation}
\begin{aligned}
\bm{T}'_{T}=\alpha \cdot \bm{Z}_{T}\bm{W}^O + \beta \cdot \bm{T}_{T}
\end{aligned}
\label{eq10}
\end{equation}
\begin{equation}
\begin{aligned}
\hat{\bm{T}}_{T}=\gamma \cdot \bm{T}'_{T} + \delta \cdot {\rm FFN}(\bm{T}'_{T})
\end{aligned}
\label{eq11}
\end{equation}

\noindent Besides, we also employ multi-head cross-attention mechanism with 8 parallel heads in this paper, which enables the model to jointly understand the correlation between RGB and thermal features from different perspectives.

Thirdly, the vector $\bm{Z}_{T}$ is reprojected back to the original space through nonlinear transformation, and added to the input sequence through a residual connection \cite{he2016deep} (Eq. \ref{eq10}), where $\bm{W}^O \in \mathbb{R}^{C \times C}$ denotes a output weight matrix before FFN layer.

Finally, the feed-forward network (FFN) with two fully-connected layers as that in the standard Transformer \cite{dosovitskiy2020image} is applied to further refine the global information to improve the robustness and accuracy of the model and output the enhanced features $\hat{\bm{T}}_{T}$ (Eq.~\ref{eq11}).

Inspired by \cite{shen2022sliced}, we apply learnable coefficients on each branch of residual connection in Eq. \ref{eq10} and Eq. \ref{eq11}, adaptively learning the data from different branches to achieve performance gain, where $\alpha, \beta, \gamma, \delta$ are the learnable parameters initialized as 1 during training.

Similar to the thermal branch, the other CFE module is also utilized to enhance the features of RGB branch, which can be formulated in Eq. \ref{eq12}:
\begin{equation}
\begin{aligned}
\hat{\bm{T}}_{R}=\mathcal{F}_{CFE-R}(\{\bm{{T}}_{R},\bm{{T}}_{T}\})
\end{aligned}
\label{eq12}
\end{equation}

It is worth to mention that CFT \cite{qingyun2021cross} is also a transformer-based method, which directly concatenates the tokens of each modality and computes the correlation across modalities with a single transformer encoder. Differently, we employ two improved cross-attention transformers to compute the correlation across modalities only with queries from auxiliary modality, which have lower computational complexity and fewer parameters.
The detailed computational complexity comparison between CFT and our method is given in Tab. \ref{tab_1}.

\begin{table}[h]
    \centering
	\caption{The computational complexity  comparison between CFT and our proposed method.($T$ is the length of tokens, while $C$ is the channels of token.)}
	\label{tab_1}
	\renewcommand\arraystretch{1.3}
	\scriptsize
    \begin{tabular}{c|c|c}
    \toprule
    \textbf{Step} & \textbf{CFT} & \textbf{Ours} \\ \midrule
    $QK^T$              & $O(4T^2 \times C)$             & $O(2T^2 \times C)$              \\ 
    $\rm{softmax(}\frac{QK^T}{\sqrt{D_K}}\rm{)}$              & $O(4T^2)$             & $O(2T^2)$              \\ 
    $\rm{softmax(}\frac{QK^T}{\sqrt{D_K}}\rm{)} \cdot V$              & $O(4T^2 \times C)$             & $O(2T^2 \times C)$              \\ 
    $\rm{FFN}$              & $O(16T \times C^2)$             & $O(8T \times C^2)$              \\ 
    $\rm{Total}$              & $O(4T^2 \times C + 16T \times C^2)$             & $O(2T^2 \times C + 16T \times C^2)$              \\ \bottomrule
    \end{tabular}
\end{table}

\subsubsection{Spatial Feature Shrinking (SFS)}
\label{section:3.2.2}
Although the initial feature maps used in fusion are downsampled using backbone, the model's parameters and memory cost can still much surpass the operating requirements of standard processors. 
To lower down the subsequent computational cost of our module with less information loss in the feature maps, we apply a SFS module before the CFE module that compresses the feature maps. 
In this module, we attempt two different methods with convolution and pooling operations, and the details are as follows.

\textbf{Convolution operation.}
We first design a method for dimension reduction based on the convolution operation, as shown in Eq \ref{eq13}. Specifically, we transform the spatial information of features to the channel dimension by reshaping the the dimensions of feature maps, and then compress the channel dimension with $1\times1$ convolution operation.
\begin{equation}
\begin{aligned}
\bm{F}_{conv}=conv_{1\times1}(Reshape(\bm{F}))
\end{aligned}
\label{eq13}
\end{equation}

\noindent where $\bm{F}$ denotes the input feature maps. $\bm{F}_{conv}$ denotes the compressed feature maps by $1\times1$ convolution.

\textbf{Pooling operation.}
Average pooling and max pooling are two conventional pooling methods, which commonly used to reduce the spatial dimension of feature maps without additional parameters. Average pooling computes the mean of all the elements in the pooling region and retains the background information in the images, while max pooling considers the maximum element in the pooling region and mainly retains the texture features of objects. Thus, we employ a method aggregating average pooling and max pooling adaptively inspired from mixed pooling \cite{yu2014mixed}, as shown in Eq. \ref{eq16}.
\begin{equation}
\begin{aligned}
\bm{F_a}=\mathrm{AvgPooling}(\bm{F},S),~ 
\bm{F}_m=\mathrm{MaxPooling}(\bm{F},S)
\end{aligned}
\label{eq14}
\end{equation}
\begin{equation}
\begin{aligned}
\bm{F}_o=\lambda \cdot \bm{F}_a+(1-\lambda) \cdot \bm{F}_m
\end{aligned}
\label{eq16}
\end{equation}

\noindent where $\bm{F}$ denotes the input feature maps. $S$ denotes scaling factors of feature maps. $\bm{F}_a$ and $\bm{F}_m$ denote the compressed feature maps via $\mathrm{AvgPooling}(\cdot)$ and  $\mathrm{MaxPooling}(\cdot)$ respectively. $\lambda$ denotes the weight between 0 and 1, which is a learnable parameter in this paper.

Compared with the original feature maps of dimension $H \times W \times C$, the compressed feature maps have dimensions $(H \times W)/S \times C$, resulting in the dimension of tokens reduced from $HW \times C$ to $HW/S \times C$. Thus, the dimensions of keys, queries and values in the CFE module become $K,Q,V \in \mathbb{R}^{HW/S \times C}$. Finally, the total computational complexity is reduced from $O(W^2H^2 \times C + 8WH \times C^2)$ to $O(W^2H^2/S^2 \times C + 8WH/S \times C^2)$.

\begin{figure*}[]
	\centering
	\setlength{\abovecaptionskip}{0.cm}
	\subfigure[]{
		\begin{minipage}[b]{1\linewidth}
			\includegraphics[width=.8\linewidth]{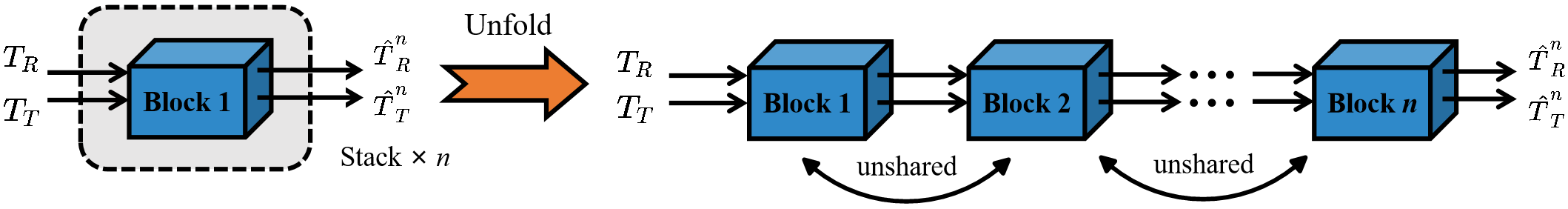}
		\end{minipage}
	}\hspace{-3 mm}	
	\subfigure[]{
		\begin{minipage}[b]{1\linewidth}
			\includegraphics[width=.8\linewidth]{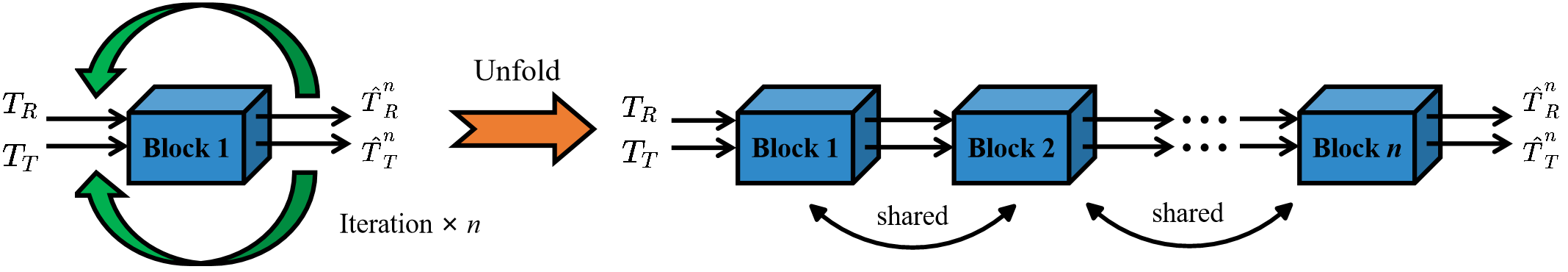}
		\end{minipage}
	}
	\caption{Visualization of difference between the traditional method and ours. (a) Traditional  method stacks multiple blocks in series, and parameters are not shared in each block. (b) Our proposed ICFE module iteratively refines the features across modalities, and  parameters are shared in each block. Block in the image denotes our proposed dual CFE modules.}
	\label{fig:Fig4}
\end{figure*}

\subsubsection{Iterative Cross-modal Feature Enhancement (ICFE)}
For the sake of strengthening the memory of complementary information from inter-modal and intra-modal features to further improve the model performance, we introduce an iterative learning strategy based on the CFE module and dubbed as ICFE module. 
As illustrated in Fig. \ref{fig:Fig4} (a), the traditional methods generally improve the performance by stacking serveral modules, however this strategy of dramatically expanding the depth of the model may not only increase the parameters significantly, but also lead to overfitting.
Instead, our proposed iterative learning strategy deepens the depth of the network over multiple iterations with shared parameters, and progressively refines the complementary information across modalities without increasing the number of parameters, as shown in Fig. \ref{fig:Fig4} (b). Taking $n$ iterations as an example, it can be simplified as follows (Eq.~\ref{eq17}):
\begin{equation}
\begin{aligned}
\{\bm{\hat{T}}^{n}_{R},\bm{\hat{T}}^n_{T}\} &= \mathcal{F}_{ICFE}(\{\bm{{T}}_{R},\bm{{T}}_{T}\}, n) \\
      &= \underbrace{\mathcal{F}_{CFE}(\cdots  \mathcal{F}_{CFE}}_{n}(\{\bm{{T}}_{R},\bm{{T}}_{T}\}))
\end{aligned}
\label{eq17}
\end{equation}

\noindent where $\{\bm{\hat{T}}^{n}_{R}, \bm{\hat{T}}^n_{T}\}$ denotes the output sequence obtained after $n$ iterative operations, $\{\bm{{T}}_{R},\bm{{T}}_{T}\}$ denotes the input sequence of the ICFE module. $\mathcal{F}_{ICFE}(\cdot)$ denotes our proposed ICFE module, which integrates two CFE modules for RGB and thermal branch respectively. 
The output of each iterative operation is used as the input of the next iterative operation, and parameters are shared between each iterative operation. Besides, The output sequences $\bm{\hat{T}}^{n}_{R}$ and $\bm{\hat{T}}^n_{T}$ from ICFE module are first converted into the feature maps, and then re-calibrated to the original size of feature map with bilinear interpolation.

\begin{figure}[]
	\centering
	\includegraphics[width=1\linewidth]{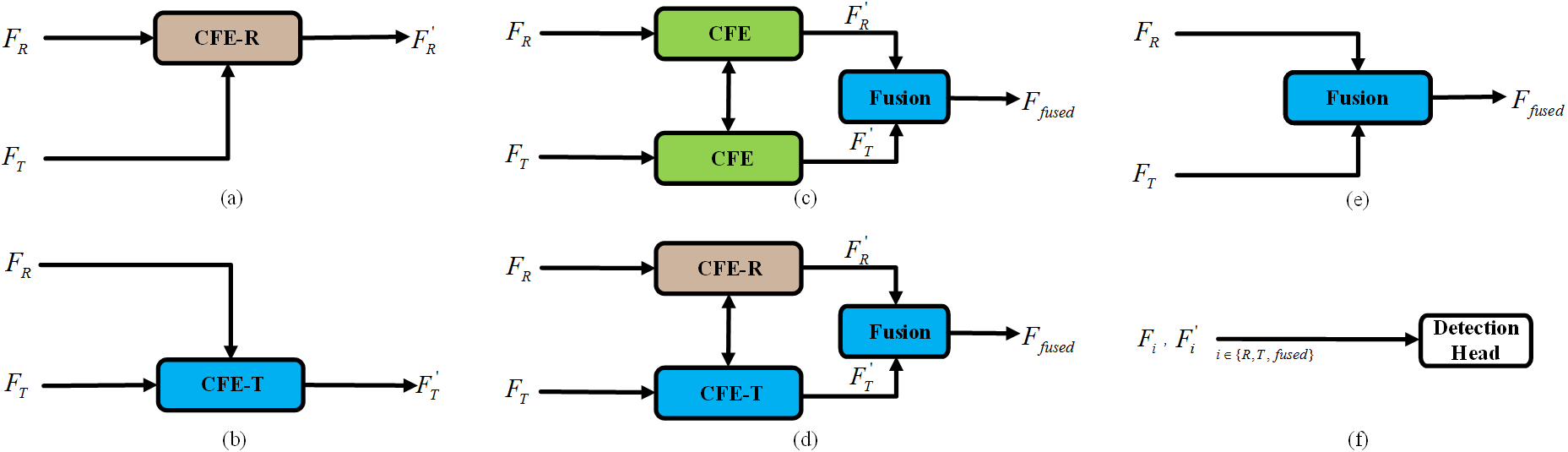}
	\caption{Different fusion modes of CFE module. (a) Single CFE module for RGB modality. (b) Single CFE module for thermal modality. (c) Dual CFE module (shared) for both RGB and thermal modality. (d) Dual CFE modules for both RGB and thermal modality. (e) Baseline feature fusion for both RGB and thermal modality. (f) Detection head from all these output features.}
	\label{fig:Fig_head}
\end{figure}

\subsubsection{Fusion Modes for Detection Heads}
Figure ~\ref{fig:Fig3} shows how our proposed CFE module can work with different input modalities. We have investigated four alternative fusion modes to validate the effectiveness of the CFE module, as shown in Fig.~\ref{fig:Fig_head}. Only a single modality feature is outputted in Fig.~\ref{fig:Fig_head} (a) and (b), forcing the CFE-R and CFE-T module to collect complimentary features from thermal and RGB image features respectively. In addition, we have also explored two different work modes with dual CFE modules, which used shared and unshared parameters, as shown in Fig. \ref{fig:Fig_head} (c) and (d). 
Fig.~\ref{fig:Fig_head} (e) illustrates the baseline feature fusion method with NIN fusion method. Finally, all these fused feature maps ($F_{i}, F^{'}_{i},i=\{R,T,fused\}$) will be fed into the detection head as shown in Fig.~\ref{fig:Fig_head} (f).
It's noteworthy to see that our method naturally favors both dual and single image modalities. Thanks to the cross-attention mechanism, even if one of the input modalities is missing or the image quality is poor, our method can still produce satisfactory results. The detailed experimental study which can support our assertion will be postponed in Section  \ref{section:4.3}.

\section{Experiments}
\subsection{Datasets and Evaluation metrics}
\textbf{KAIST Dataset.} KAIST \cite{hwang2015multispectral} is a popular multispectral pedestrian detection benchmark, which involves scenes with different illuminations. There are 8,963 and 2,252 weakly-aligned image pairs with the resolution of $640\times512$ for training \cite{zhou2020improving} and testing respectively. The performance evaluation on KAIST dataset is typically in accordance with the metric log-average miss rate \cite{dollar2011pedestrian}. For more accurate annotations, we use the sanitized annotations for training \cite{zhang2019weakly} and testing \cite{liu2016multispectral}.

\textbf{FLIR Dataset.} FLIR \cite{flir_dataset} is a challenging multispectral object detection dataset including daytime and night scenes. There are 5,142 aligned multispectral image pairs, of which 4,129 are used for training and 1,013 for testing. It contains three classes of objects, namely "person", "car" and "bicycle". Since the images are misaligned in the original dataset,  we select the FLIR-aligned version \cite{zhang2020multispectral} for comparisons in our experiment. 

\textbf{VEDAI Dataset.} VEDAI \cite{razakarivony2016vehicle} is a public dataset for small target detection in aerial imagery, which contains more than 3,700 annotated targets in 1,268 RGB-infrared image pairs. There are 9 vehicle categories in this dataset. We use the images with size of $1024\times1024$ for training and testing, and convert the annotations to the horizontal-box format with \cite{qingyun2022cross} as reference since the original version is annotated as a rotating box with four-corner coordinates.

\textbf{Log-average Miss Rate.} The log-average miss rate ($MR^{-2}$) \cite{dollar2011pedestrian} is used for the evaluation on KAIST dataset. It represents the average miss rate under 9 FPPI values, which are sampled uniformly in the logarithmic interval [$10^{-2}$,1]. The lower values of $MR^{-2}$, the better performance.

\textbf{Average Precision.} Average Precision (AP) is a common evaluation metric for object detection. The positive and negative samples should be divided according to the the correctness of classification and Intersection over Union (IoU) threshold. Usually, 0.5 is used as the IoU threshold. In general, mean Average Precision (mAP) represents the average of AP under all categories. Different from $MR^{-2}$, the higher values of AP and mAP, the better performance.

\subsection{Implementation Details}
Our method is implemented using PyTorch 1.7.1 framework on a Ubuntu 18.04 server with CPU i7-9700, 64G Memory and Nvidia RTX 3090 24G GPU. The training phase takes 60 epochs with the batch size of 8. The SGD optimizer is used with the initial learning rate of $1.0\times10^{-2}$ and the momentum of 0.937. In addition, the weight decay factor is 0.0005 and the learning rate decay method is cosine annealing. The input size of images are $640\times640$ for training and $640\times512$ for testing. Besides, mosaic and random flipping is used for data augmentation. The loss function is utilized following the detectors of YOLOv5 and FCOS in the orignal paper. In the ablation studies, we use YOLOv5 with NIN fusion \cite{limultispectral} as default baseline for comparison.

\subsection{Ablation Study}
\label{section:4.3}

\subsubsection{Effects of learnable parameters applied on residual connection}
Given that Shen et al. \cite{shen2022sliced} has proved learning parameters applied on both branches is slightly better than that on the single branch, we evaluate the effectiveness of learnable parameters applied on both branches of residual connection in our proposed CFE module. The experimental results are given in Table. \ref{tab_2}. Compared with the CFE module without learnable parameters, adding learnable parameters on both branches reduces MR from 7.86$\%$ to 7.63$\%$ on the KAIST dataset, and improves mAP50 from 77.1$\%$ to 77.5$\%$ on the FLIR dataset. As a result, the learnable parameters applied on both branches of residual connection are effective to achieve performance gain without significantly increasing the computational cost in our CFE module.

\begin{table}[h]
    \centering
	\caption{Effects of learnable parameters in CFE module on both KAIST and FLIR datasets. (LP denotes learning parameters applied on both branches.)}
	\label{tab_2}
	\renewcommand\arraystretch{1.1}
	\scriptsize
    \begin{tabular}{c|c|c|c|c}
    \toprule
    \multirow{2}{*}{\textbf{Method}} & \multirow{2}{*}{\textbf{LP}}                  & \textbf{KAIST}                                             & \textbf{FLIR}                                             & \multirow{2}{*}{\textbf{Params(M)}}                   \\ \cline{3-4}
                                     &                                               & \textbf{MR(\%)$\downarrow$}                                            & \textbf{mAP50(\%)$\uparrow$}                                        &                                                       \\ \midrule
    Baseline+CFE                     & \begin{tabular}[c]{@{}c@{}}\\ \checkmark\end{tabular} & \begin{tabular}[c]{@{}l@{}}7.86\\ 7.63(-0.23)\end{tabular} & \begin{tabular}[c]{@{}l@{}}77.1\\ 77.5(+0.4)\end{tabular} & \begin{tabular}[c]{@{}c@{}}120.2\\ 120.2\end{tabular} \\ \bottomrule
    \end{tabular}
\end{table}

\subsubsection{Effects of CFE module for mono-modality and dual-modality}
The experimental results of CFE module for mono-modality on both KAIST and FLIR datasets are presented in Tab. \ref{tab_3}, which is separated into three groups acoording to the output modality. 
In the first group (first row), we apply the CFE module to enhance RGB features leveraging the compensatory information from the thermal images and only output the enhanced RGB features for the subsequent detection, as shown in Fig. \ref{fig:Fig_head}(a). Our dual branch method with CFE module outperforms the RGB-only single branch method by 0.65$\%$ and 0.90$\%$ on the KAIST and FLIR dataset respectively. Similarly, in the second group (second row), the enhanced thermal feature with CFE module in Fig. \ref{fig:Fig_head}(b) achieves a gain of 0.59$\%$ and 1.20$\%$ over the thermal-only detector on the KAIST and FLIR dataset respectively. The top two rows in Tab. \ref{tab_3} indicate that the quality of RGB features on the KAIST dataset is superior to that of thermal features for detection, whereas thermal features on the FLIR dataset are superior to RGB features in quality. This might be caused by the properties of dataset, the camera model and other elements.
Thus, the dual CFE modules are applied to RGB and thermal branch to collect the complementary information from each other, and fused features from enhanced RGB and thermal modality are utilized for the subsequent detection in the last group (last row), which outperforms the baseline method by 0.70$\%$ and 1.00$\%$ on the KAIST and FLIR dataset respectively.
As a result, the experimental findings presented above demonstrate the effectiveness of our proposed CFE module, which favors both RGB and thermal-based global feature fusion.

\begin{table}[]
    \centering
	\caption{Effects of CFE module for each modality on both KAIST and FLIR datasets.(The lower the MR, the better. The higher mAP, the better performance. In the third column, the letter (a)$\sim$(f) denote the fusion mode in Fig.~\ref{fig:Fig_head})}
	\label{tab_3}
	\renewcommand\arraystretch{1}
	\scriptsize
    \begin{tabular}{l|l|l|l|l}
    \toprule
    \multicolumn{1}{c|}{\multirow{2}{*}{\textbf{Input Modality}}}                   & \multicolumn{1}{c|}{\multirow{2}{*}{\textbf{Output Feature}}}                   & \multicolumn{1}{c|}{\multirow{2}{*}{\textbf{Method}}}                                      & \multicolumn{1}{c|}{\textbf{KAIST}}                                       & \multicolumn{1}{c}{\textbf{FLIR}}                                      \\ \cline{4-5} 
    \multicolumn{1}{c|}{}                                                           & \multicolumn{1}{c|}{}                                                           & \multicolumn{1}{c|}{}                                                                      & \multicolumn{1}{c|}{\textbf{MR(\%)$\downarrow$}}                                       & \multicolumn{1}{c}{\textbf{mAP50(\%)$\uparrow$}}                                  \\ \midrule
    \begin{tabular}[c]{@{}l@{}}RGB\\ RGB+Thermal\end{tabular}                       & \begin{tabular}[c]{@{}l@{}}$F_R$\\ $F_R$\end{tabular}                           & \begin{tabular}[c]{@{}l@{}}Baseline-RGB (f)\\ Baseline+CFE (a)\end{tabular}                & \begin{tabular}[c]{@{}l@{}}18.39\\ 17.74(-0.65)\end{tabular}              & \begin{tabular}[c]{@{}l@{}}67.8\\ 68.7(+0.9)\end{tabular}              \\ \midrule \midrule
    \begin{tabular}[c]{@{}l@{}}Thermal\\ RGB+Thermal\end{tabular}                   & \begin{tabular}[c]{@{}l@{}}$F_T$\\ $F_T$\end{tabular}                           & \begin{tabular}[c]{@{}l@{}}Baseline-Thermal (f)\\ Baseline+CFE (b)\end{tabular}            & \begin{tabular}[c]{@{}l@{}}18.94\\ 18.35(-0.59)\end{tabular}              & \begin{tabular}[c]{@{}l@{}}73.9\\ 75.1(+1.2)\end{tabular}              \\ \midrule \midrule
    \begin{tabular}[c]{@{}l@{}}RGB+Thermal\\ RGB+Thermal\\ RGB+Thermal\end{tabular} & \begin{tabular}[c]{@{}l@{}}$F_{fused}$\\ $F_{fused}$\\ $F_{fused}$\end{tabular} & \begin{tabular}[c]{@{}l@{}}Baseline (e)\\ Baseline+CFE (c)\\ Baseline+CFE (d)\end{tabular} & \begin{tabular}[c]{@{}l@{}}8.33\\ 10.78(+2.45)\\ 7.63(-0.70)\end{tabular} & \begin{tabular}[c]{@{}l@{}}76.5\\ 76.0(-0.5)\\ 77.5(+1.0)\end{tabular} \\ \bottomrule
    \end{tabular}
\end{table}

\subsubsection{Effects of the number of stacked modules}
\begin{figure}[h]
	\centering
	\includegraphics[width=1\linewidth]{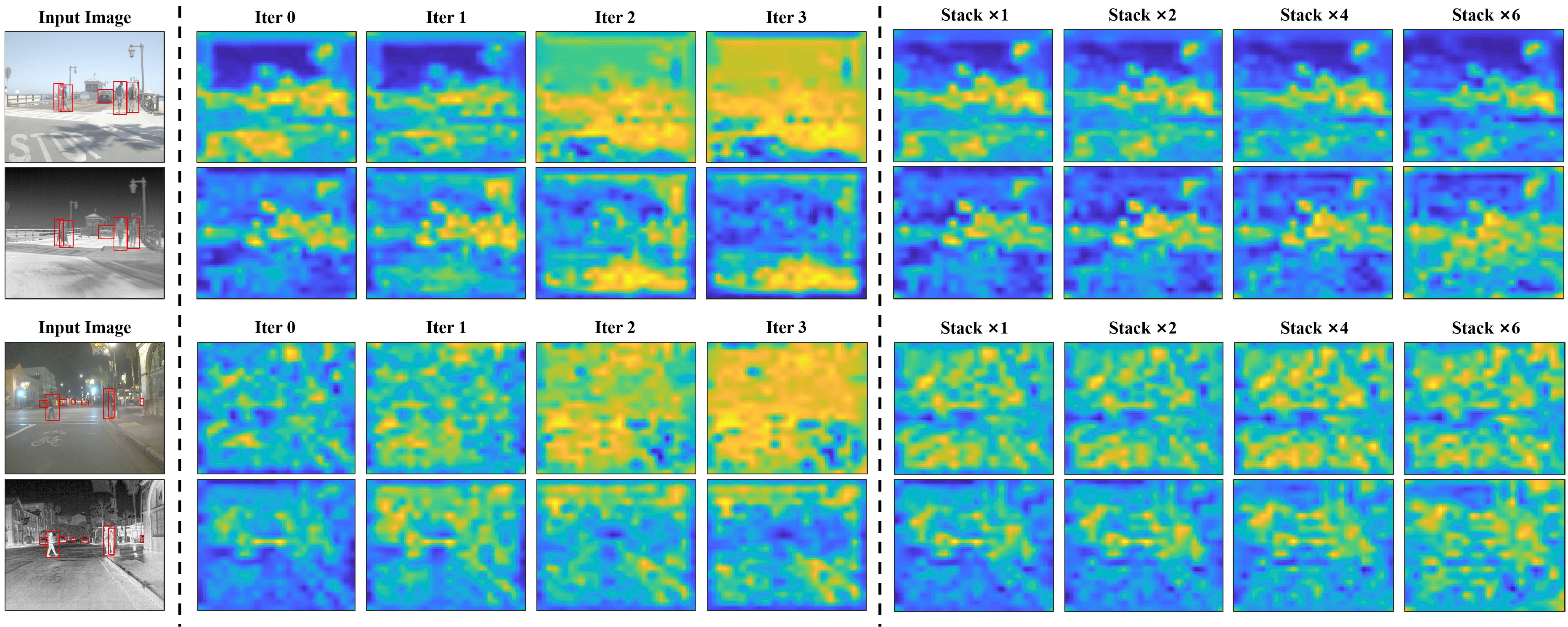}
	\caption{Visualization results of CFE module with different stacking number and ICFE module with different iteration number on the FLIR dataset. The top row is an image pair at daytime, and the bottom row is an example at night. The 1st column is the input image; the $2^{nd}\sim 5^{th}$ columns are feature maps from iterative learning; the $6^{th} \sim 9^{th}$ columns are feature maps from different numbers of stacking.}
	\label{fig:iter_stack}
\end{figure}

In this section, we provide the mAP values for different number of stacked CFE modules on FLIR dataset. Tab. 4 shows that as the number of stacked modules increases to 10, the parameter numbers and GPU memory increases by more than $4\times$ times, while the running speed decreases dramatically from 40.5 Hz to 17.3 Hz with a marginal benefit of 0.70$\%$ in terms of mAP. The previous studies \cite{raghu2021vision, venkataramanan2023skip} find that the attention maps across adjacent layers of vision transformer exhibit very high similarity. 
As shown in Fig. \ref{fig:iter_stack}(right), after visualizing the feature maps of different stacked numbers, we also find this phenomenon in our experiment.
So, we think that the high similarity of feature maps can result in marginal performance improvement. As a result, stacking blocks in series is not an efficient solution for feature fusion.

\begin{table}[]
    \centering
	\caption{Comparison with different number of stacked modules on the FLIR dataset.}
	\label{tab_4}
	\renewcommand\arraystretch{1}
	\scriptsize
    \begin{tabular}{c|c|c|c|c}
    \toprule
    \textbf{Number}                                                & \textbf{mAP50(\%)$\uparrow$}                                                               & \textbf{Mem(M)}                                                                                                & \textbf{Params(M)}                                                                    & \textbf{FPS(Hz)}                                                                \\ \midrule
    \begin{tabular}[c]{@{}c@{}}1\\ 2\\ 4\\ 6\\ 8\\ 10\end{tabular} & \begin{tabular}[c]{@{}c@{}}77.5\\ 77.6\\ 77.4\\ 77.8\\ 77.9\\ 78.2\end{tabular} & \multicolumn{1}{r|}{\begin{tabular}[c]{@{}r@{}}528.5\\ 686.0\\ 1037.5\\ 1795.5\\ 1747.0\\ 2100.5\end{tabular}} & \begin{tabular}[c]{@{}c@{}}120.2\\ 164.3\\ 252.5\\ 340.7\\ 428.9\\ 517.1\end{tabular} & \begin{tabular}[c]{@{}c@{}}40.5\\ 31.9\\ 26.9\\ 22.9\\ 19.6\\ 17.3\end{tabular} \\ \bottomrule
    \end{tabular}
\end{table}

\subsubsection{Effects of different number of iterations}
\label{section:4.4.3}
The experimental results of varying number of iterations on the KAIST and FLIR datasets are shown in Tab. \ref{tab_5}. With only one iteration, the iterative learning method reduces MR from 7.63$\%$ to 7.17$\%$ on the KAIST dataset and improves mAP50 from 77.50$\%$ to 79.20$\%$ on the FLIR dataset. Interestingly, we find that extra iterations do not boost performance, and one iteration achieves the best results in our experiment. As shown in Fig. \ref{fig:iter_stack}(middle), we also visualize the feature maps of ICFE modules. As the number of iterations rises, we find that the interaction between different modality features results in negative effects and the background information are gradually enhanced. We think that the enhanced interference of background noises may lead to performance degradation. Furthermore, since the iterative learning technique uses shared parameters, more iterations will not incur extra parameters or memory costs. 

Comparing the experimental results in Tab. \ref{tab_4} and Tab. \ref{tab_5}, we can infer that interative learning is more effective for cross-modal feature interaction and also outperforms the stacking method. Besides, our method has a much faster inference speed of 36.7 FPS than the stacking method.

\begin{table}[]
    \centering
    \caption{Comparison with different number of iterations.}
	\label{tab_5}
	\renewcommand\arraystretch{1}
	\scriptsize
    \begin{tabular}{c|c|c|c|c|c}
    \toprule
    \multirow{2}{*}{\textbf{Number}} & \textbf{KAIST}  & \textbf{FLIR}      & \multirow{2}{*}{\textbf{Params(M)}} & \multirow{2}{*}{\textbf{Mem(M)}} & \multirow{2}{*}{\textbf{FPS(Hz)}} \\ \cline{2-3}
                                     & \textbf{MR(\%)} & \textbf{mAP50(\%)} &                                     &                                  &                                   \\ \midrule
    0                                & 7.63            & 77.5               & 120.2                               & 528.5                            & 40.5                              \\
    1                                & \textbf{7.17}   & \textbf{79.2}      & 120.2                               & 528.5                            & 36.7                              \\
    2                                & 7.87            & 76.9               & 120.2                               & 528.5                            & 32.8                              \\
    3                                & 7.87            & 77.5               & 120.2                               & 528.5                            & 29.6                              \\ \bottomrule
    \end{tabular}
\end{table}

\subsubsection{Effects of different spatial feature shrinking methods}
We evaluate multiple existing approaches in order to find a reliable down-sampling method with less loss of feature information, and the experiment results are shown in Tab. \ref{tab_6}. In comparison to the other down-sampling approaches, mixed pooling  produces best results, with MR of $7.17\%$ and mAP50 of $79.20\%$ on the KAIST and FLIR datasets, respectively. As a result, we use mixed pooling to compress the feature maps and reduce computational complexity in this paper.

\begin{table}[h]
	\centering
	\caption{Comparison with different spatial feature shrinking methods.}
	\label{tab_6}
	\renewcommand\arraystretch{1}
	\scriptsize
	\begin{tabular}{c|c|c}
		\toprule
		\multirow{2}{*}{\textbf{Methods}} & \textbf{KAIST}  & \textbf{FLIR}      \\ \cline{2-3} 
		& \textbf{MR(\%)} & \textbf{mAP50(\%)} \\ \midrule
		Average Pooling                   & 7.58            & 77.0               \\
		Max Pooling                       & 7.94            & 78.4               \\
		Ours-Conv                         & 7.42            & 78.6               \\
		Ours-Pool                         & 7.17            & 79.2               \\ \bottomrule
	\end{tabular}
\end{table}

\subsubsection{Discussion on Different Input Modalities}
In this section, we have also conducted four groups of experiments in order to validate the effectiveness of using different input modalities for CFE, and the experimental results are shown in Tab. \ref{tab_8}. The first group (row $1\sim2$) demonstrates the experimental result in the KAIST and FLIR datasets for the YOLOv5 detector with a single input image modality (RGB or thermal). In the second group (row $3\sim5$), we provide the results of the YOLOv5+NIN method with one or two input image modalities (RGB or RGB+Thermal). It is clear to observe that the performance of the YOLOv5+NIN method with two different input modalities (row 3) is superior to the YOLOv5 method (row 1 and 2) by a large margin. However, the YOLOv5+NIN method will bring a large performance degradation when using the same two modalities (row $4\sim5$) as input. Furthermore, we also conducts experiments with our proposed method (YOLOv5+ICFE) in the third group (row $6\sim9$). It is surprisingly found that using the same two modality images can still achieve competitive results comparing to the those with both RGB and thermal images. It indicates that our method can provide discriminative mono-modal features for the subsequent detection stage with a slight performance degradation as shown in row 7 and 9 on both KAIST and FLIR datasets. This is beneficial to the scenarios where one of the input modalities is missing or the image quality is poor. 
In the last group (row $10\sim12$), we also have observed significant drop in the detection performance of our proposed method (YOLOv5+ICFE+NIN) due to the append of the NIN module to the output of the dual CFE modules. The observation from both row $4\sim5$ and $11\sim12$ indicates that NIN is harmful to both the YOLOv5+NIN and our proposed method when only one input modality is accessible.

\begin{table}[h]
	\centering
	\caption{Comparison with different input modalities. (R denotes RGB, T denotes Thermal. R+T represents the input with dual modalities, while R+R or T+T denotes input with single modality and ignores the other modality. In the  third column, the letter (a)$\sim$(f) denote the fusion mode in Fig.~\ref{fig:Fig_head})}
	\label{tab_8}
	\renewcommand\arraystretch{1}
	\setlength{\tabcolsep}{3pt}
	\scriptsize
	\begin{tabular}{c|c|l|l|l|l}
		\toprule
		\multirow{2}{*}{\textbf{Number}}                                       & \multirow{2}{*}{\textbf{Methods}} & \multicolumn{1}{c|}{\multirow{2}{*}{\textbf{Input  }}}                                  & \multicolumn{1}{c|}{\multirow{2}{*}{\textbf{Output }}}                                    & \multicolumn{1}{c|}{\textbf{KAIST}}                                                                  & \multicolumn{1}{c}{\textbf{FLIR}}                                                              \\ \cline{5-6} 
		&                                   & \multicolumn{1}{c|}{}                                                                          & \multicolumn{1}{c|}{}                                                                            & \multicolumn{1}{c|}{\textbf{MR(\%)$\downarrow$}}                                                      & \multicolumn{1}{c}{\textbf{mAP50(\%)$\uparrow$}}                                                \\ \midrule
		\multirow{2}{*}{\begin{tabular}[c]{@{}c@{}}1\\ 2\end{tabular}}         & \multirow{2}{*}{YOLOv5}           & \multirow{2}{*}{\begin{tabular}[c]{@{}l@{}}R (f)\\ T (f)\end{tabular}}                         & \multirow{2}{*}{\begin{tabular}[c]{@{}l@{}}$F_R$\\ $F_T$\end{tabular}}                           & \multirow{2}{*}{\begin{tabular}[c]{@{}l@{}}18.39\\ 18.94\end{tabular}}                               & \multirow{2}{*}{\begin{tabular}[c]{@{}l@{}}67.8\\ 73.9\end{tabular}}                           \\
		&                                   &                                                                                                &                                                                                                  &                                                                                                      &                                                                                                \\ \midrule
		\multirow{3}{*}{\begin{tabular}[c]{@{}c@{}}3\\ 4\\ 5\end{tabular}}     & \multirow{3}{*}{YOLOv5+NIN}       & \multirow{3}{*}{\begin{tabular}[c]{@{}l@{}}R+T (e)\\ R+R (e)\\ T+T (e)\end{tabular}}           & \multirow{3}{*}{\begin{tabular}[c]{@{}l@{}}$F_{fused}$\\ $F_{fused}$\\ $F_{fused}$\end{tabular}} & \multirow{3}{*}{\begin{tabular}[c]{@{}l@{}}8.33\\ 43.79(+35.46)\\ 43.79(+35.46)\end{tabular}}        & \multirow{3}{*}{\begin{tabular}[c]{@{}l@{}}76.5\\ 57.1(-19.4)\\ 65.3(-11.2)\end{tabular}}      \\
		&                                   &                                                                                                &                                                                                                  &                                                                                                      &                                                                                                \\
		&                                   &                                                                                                &                                                                                                  &                                                                                                      &                                                                                                \\ \midrule
		\multirow{4}{*}{\begin{tabular}[c]{@{}c@{}}6\\ 7\\ 8\\ 9\end{tabular}} & \multirow{4}{*}{YOLOv5+ICFE}      & \multirow{4}{*}{\begin{tabular}[c]{@{}l@{}}R+T (a)\\ R+R (a)\\ R+T (b)\\ T+T (b)\end{tabular}} & \multirow{4}{*}{\begin{tabular}[c]{@{}l@{}}$F_R$\\ $F_R$\\ $F_T$\\ $F_T$\end{tabular}}           & \multirow{4}{*}{\begin{tabular}[c]{@{}l@{}}17.74\\ 20.50(+2.76)\\ 18.35\\ 20.07(+1.72)\end{tabular}} & \multirow{4}{*}{\begin{tabular}[c]{@{}l@{}}68.7\\ 66.3(-2.4)\\ 75.1\\ 74.2(-0.9)\end{tabular}} \\
		&                                   &                                                                                                &                                                                                                  &                                                                                                      &                                                                                                \\
		&                                   &                                                                                                &                                                                                                  &                                                                                                      &                                                                                                \\
		&                                   &                                                                                                &                                                                                                  &                                                                                                      &                                                                                                \\ \midrule
		\multirow{3}{*}{\begin{tabular}[c]{@{}c@{}}10\\ 11\\ 12\end{tabular}}  & \multirow{3}{*}{YOLOv5+ICFE+NIN}  & \multirow{3}{*}{\begin{tabular}[c]{@{}l@{}}R+T (d)\\ R+R (d)\\ T+T (d)\end{tabular}}           & \multirow{3}{*}{\begin{tabular}[c]{@{}l@{}}$F_{fused}$\\ $F_{fused}$\\ $F_{fused}$\end{tabular}} & \multirow{3}{*}{\begin{tabular}[c]{@{}l@{}}7.17\\ 31.23(+23.06)\\ 37.42(+29.79)\end{tabular}}        & \multirow{3}{*}{\begin{tabular}[c]{@{}l@{}}79.2\\ 57.8(-21.4)\\ 66.0(-13.2)\end{tabular}}      \\
		&                                   &                                                                                                &                                                                                                  &                                                                                                      &                                                                                                \\
		&                                   &                                                                                                &                                                                                                  &                                                                                                      &                                                                                                \\ \bottomrule
	\end{tabular}
\end{table}

\subsubsection{Comparisons with different backbones and heads}
In order to evaluate the effectiveness and generality of our proposed DMFF module, we first conduct the experiments on YOLOv5 detector with three different backbones: VGG16, ResNet50 and CSPDarkNet53. 
As shown in Tab. \ref{tab_7}, the results on the KAIST dataset show that our approach outperforms the baseline method by 0.66$\%$, 0.97$\%$ and 1.16$\%$ on VGG16, ResNet50 and CSPDarknet53 respectively. 
The results on the FLIR dataset demonstrate that our approach also achieves a gain of 0.50$\%$, 1.50$\%$ and 2.70$\%$ over the baseline method on VGG16, ResNet50 and CSPDarknet53 respectively. 
As a result, we conclude that our proposed DMFF module is applicable to various backbones and is effective under different evaluation metrics.

We also evaluate on the FCOS detector to further examine the effectiveness and generality of our proposed DMFF module. 
The experimental results are given in Tab. \ref{tab_7}. 
Compared with the baseline method, FCOS with DMFF module reduces MR from 14.03$\%$ to 12.96$\%$ with a gain of 1.07$\%$ on the KAIST dataset, and improves mAP50 from 69.80$\%$ to 71.70$\%$ with a gain of 1.90$\%$ on the FLIR dataset. The above results indicate that our proposed DMFF module works wells for both anchor-base and anchor-free detectors. 
Finally, it is clear to find that YOLOv5 detector with CSPDarknet53 backbone achieves the best performance with a fair number of parameters when compared to the other backbones and detectors.

\begin{table}[h]
    \centering
	\caption{Comparison with different detectors and backbones.}
	\label{tab_7}
	\renewcommand\arraystretch{1}
	\setlength{\tabcolsep}{3pt}
	\scriptsize
    \begin{tabular}{c|c|l|c|c|c|c|c}
    \toprule
    \multirow{2}{*}{\textbf{Detector}} & \multirow{2}{*}{\textbf{Type}}   & \multirow{2}{*}{\textbf{Backbone}} & \multirow{2}{*}{\textbf{Method}}                        & \textbf{KAIST}                                               & \textbf{FLIR}                                             & \multirow{2}{*}{\textbf{Params(M)}}           & \multirow{2}{*}{\textbf{FPS(Hz)}}             \\ \cline{5-6}
                                       &                                  &                                    &                                                         & \textbf{MR(\%)$\downarrow$}                                              & \textbf{mAP50(\%)$\uparrow$}                                         &                                               &                                               \\ \midrule
    \multirow{5}{*}{YOLOv5}            & \multirow{5}{*}{Anchor-based}     & VGG16                              & \begin{tabular}[c]{@{}l@{}}Baseline\\ Ours\end{tabular} & \begin{tabular}[c]{@{}l@{}}16.12\\ 15.46(-0.66)\end{tabular} & \begin{tabular}[c]{@{}l@{}}69.3\\ 69.8(+0.5)\end{tabular} & \begin{tabular}[c]{@{}l@{}}42.67\\ 62.17\end{tabular} & \begin{tabular}[c]{@{}l@{}}72.67\\ 54.48\end{tabular} \\ \cline{3-8} 
                                       &                                  & ResNet50                           & \begin{tabular}[c]{@{}l@{}}Baseline\\ Ours\end{tabular} & \begin{tabular}[c]{@{}l@{}}13.65\\ 12.68(-0.97)\end{tabular} & \begin{tabular}[c]{@{}l@{}}70.5\\ 72.0(+1.5)\end{tabular} & \begin{tabular}[c]{@{}l@{}}136.13\\ 313.76\end{tabular} & \begin{tabular}[c]{@{}l@{}}31.38\\ 31.44\end{tabular} \\ \cline{3-8} 
                                       &                                  & CSPDarkNet53                       & \begin{tabular}[c]{@{}l@{}}Baseline\\ Ours\end{tabular} & \begin{tabular}[c]{@{}l@{}}8.33\\ 7.17(-1.16)\end{tabular}   & \begin{tabular}[c]{@{}l@{}}76.5\\ 79.2(+2.7)\end{tabular} & \begin{tabular}[c]{@{}l@{}}75.44\\ 120.21\end{tabular} & \begin{tabular}[c]{@{}l@{}}50.00\\ 38.46\end{tabular} \\ \midrule \midrule
    FCOS                               & \multicolumn{1}{l|}{Anchor-free} & ResNet50                           & \begin{tabular}[c]{@{}l@{}}Baseline\\ Ours\end{tabular} & \begin{tabular}[c]{@{}l@{}}14.03\\ 12.96(-1.07)\end{tabular} & \begin{tabular}[c]{@{}l@{}}69.8\\ 71.7(+1.9)\end{tabular} & \begin{tabular}[c]{@{}l@{}}58.86\\ 65.24\end{tabular} & \begin{tabular}[c]{@{}l@{}}19.21\\ 17.32\end{tabular} \\ \bottomrule
    \end{tabular}
\end{table}

\subsection{Comparison with State-of-the-art Methods}
\textbf{KAIST Dataset.}
Tab. \ref{tab_kaist} shows the comparison of our method with existing methods on the KAIST dataset. It can be observed that our approach surpasses most of the state-of-the-art methods in the settings of reasonable, and obtains the lowest miss rate under daytime subset. 
Furthermore, Tab. \ref{tab_kaist} also illustrates that our approach runs at 38.46 Hz on the RTX 3090 platform. As a result, our method is beneficial for scenarios of object detection where high detection speed is required.

\begin{table}[]
	\centering
	\caption{Comparsion on the KAIST dataset. (Bold numbers represent the best result in each column. Methods with suffix $\dagger$ and suffix $\ddagger$ use ResNet50 and CSPDarkNet53 backbone respectively, while the others use VGG16 as defaults.)}
	\label{tab_kaist}
	\renewcommand\arraystretch{1}
	\scriptsize
	\begin{tabular}{l|llr|r|l}
		\toprule
		\multicolumn{1}{c|}{\multirow{2}{*}{\textbf{Method}}} & \multicolumn{3}{c|}{\textbf{Miss Rate(\%)$\downarrow$}}                                                                  & \multicolumn{1}{c|}{\multirow{2}{*}{\textbf{FPS(Hz)}}} & \multicolumn{1}{c}{\multirow{2}{*}{\textbf{Platform}}} \\ \cline{2-4}
		\multicolumn{1}{c|}{}                                 & \multicolumn{1}{c|}{\textbf{All}} & \multicolumn{1}{c|}{\textbf{Day}} & \multicolumn{1}{c|}{\textbf{Night}} & \multicolumn{1}{c|}{}                                  & \multicolumn{1}{c}{}                                   \\ \midrule
		ACF+T+THOG \cite{hwang2015multispectral}                                           & \multicolumn{1}{l|}{47.32}        & \multicolumn{1}{l|}{42.65}        & 56.18                               & -                                                      & -                                                      \\ 
		AR-CNN \cite{zhang2019weakly}                                               & \multicolumn{1}{l|}{10.22}        & \multicolumn{1}{l|}{10.80}        & 9.02                                & 8.33                                                   & TITAN X                                                \\ 
		CIANet \cite{zhang2019cross}                                               & \multicolumn{1}{l|}{14.13}        & \multicolumn{1}{l|}{14.78}        & 11.14                               & 16.67                                                  & GTX 1080Ti                                             \\ 
		FusionRPN+BF \cite{konig2017fully}                                         & \multicolumn{1}{l|}{18.29}        & \multicolumn{1}{l|}{19.57}        & 16.27                               & -                                                      & -                                                      \\ 
		HalfwayFusion \cite{liu2016multispectral}                                        & \multicolumn{1}{l|}{25.77}        & \multicolumn{1}{l|}{24.91}        & 26.67                               & 2.33                                                   & TITAN X                                                \\ 
		IAF-RCNN \cite{li2019illumination}                                             & \multicolumn{1}{l|}{15.57}        & \multicolumn{1}{l|}{14.81}        & 16.70                               & 4.76                                                   & TITAN X                                                \\ 
		IATDNN-IAMSS \cite{guan2019fusion}                                         & \multicolumn{1}{l|}{14.46}        & \multicolumn{1}{l|}{14.18}        & 15.28                               & 4.00                                                   & TITAN X                                                \\ 
		MBNet \cite{zhou2020improving}$\dagger$                                                & \multicolumn{1}{l|}{8.40}         & \multicolumn{1}{l|}{8.62}         & 8.27                                & 14.29                                                  & GTX 1080Ti                                             \\ 
		MLPD \cite{kim2021mlpd}                                                 & \multicolumn{1}{l|}{7.58}         & \multicolumn{1}{l|}{7.96}         & 6.95                                & -                                                      & -                                                      \\ 
		MSDS-RCNN \cite{limultispectral}                                            & \multicolumn{1}{l|}{8.23}         & \multicolumn{1}{l|}{8.83}         & \textbf{6.75}                                & 4.55                                                   & GTX 1080Ti                                             \\ 
		Ours$\ddagger$                                                  & \multicolumn{1}{l|}{\textbf{7.17}}         & \multicolumn{1}{l|}{\textbf{6.82}}         & 7.85                                & \textbf{38.46}                                                  & RTX 3090                                               \\ \bottomrule
	\end{tabular}
\end{table}

\textbf{FLIR Dataset.}
Tab. \ref{tab_flir} shows the comparison of our method with existing methods on the FLIR dataset. It is clear to observe that our approach outperforms all the existing methods and achieves state-of-the-art performance. Specifically, our method achieves 79.20$\%$, 36.9$\%$ and 41.4$\%$ in terms of mAP50, mAP75 and mAP metrics. Furthermore, our method achieves 66.90$\%$, 89.00$\%$ and 81.60$\%$ for the categories of Bicycle, Car and Person respectively.
Besides, we also employ the CFT baseline with our proposed modules (Ours*) for a fair comparison. It is clear to see that our method outperforms Ours* in terms of all mAP50, mAP75 and mAP metrics.

\begin{table}[]
    \centering
	\caption{{Comparison on the FLIR Dataset.}}
	\label{tab_flir}
	\renewcommand\arraystretch{1}
	\setlength{\tabcolsep}{3pt}
	\scriptsize
    \begin{tabular}{l|ccc|c|c|c}
    \toprule
    \multicolumn{1}{c|}{\multirow{2}{*}{\textbf{Method}}} & \multicolumn{3}{c|}{\textbf{AP50(\%)}}                                                      & \multirow{2}{*}{\textbf{mAP50(\%)}} & \multirow{2}{*}{\textbf{mAP75(\%)}} & \multirow{2}{*}{\textbf{mAP(\%)}} \\ \cline{2-4}
    \multicolumn{1}{c|}{}                                 & \multicolumn{1}{c|}{\textbf{Bicycle}} & \multicolumn{1}{c|}{\textbf{Car}} & \textbf{Person} &                                     &                                     &                                   \\ \midrule
    MMTOD-CG \cite{devaguptapu2019borrow}                                             & \multicolumn{1}{c|}{50.26}            & \multicolumn{1}{c|}{70.63}        & 63.31           & 61.4                               & -                                   & -                                 \\
    MMTOD-UNIT \cite{devaguptapu2019borrow}                                           & \multicolumn{1}{c|}{49.43}            & \multicolumn{1}{c|}{70.72}        & 64.47           & 61.5                               & -                                   & -                                 \\
    GAFF \cite{zhang2021guided}                                                 & \multicolumn{1}{c|}{-}                & \multicolumn{1}{c|}{-}            & -               & 72.9                               & 30.9                                & 37.3                              \\
    CFR \cite{zhang2020multispectral}                                                  & \multicolumn{1}{c|}{57.77}            & \multicolumn{1}{c|}{84.91}        & 74.49           & 72.4                               & -                                   & -                                 \\
    BU-ATT \cite{kieu2021bottom}$\dagger$                                               & \multicolumn{1}{c|}{56.10}            & \multicolumn{1}{c|}{87.00}        & 76.10           & 73.1                               & -                                   & -                                 \\
    BU-LTT \cite{kieu2021bottom}$\dagger$                                               & \multicolumn{1}{c|}{57.40}            & \multicolumn{1}{c|}{86.50}        & 75.60           & 73.2                               & -                                   & -                                 \\
    CFT \cite{qingyun2021cross}$\ddagger$                                                  & \multicolumn{1}{c|}{61.40}            & \multicolumn{1}{c|}{{89.50}}        & \textbf{84.10}           & 78.3                               & 35.5                                & 40.2                              \\
    Ours*$\ddagger$                                                  & \multicolumn{1}{c|}{{65.20}}            & \multicolumn{1}{c|}{\textbf{90.20}}        & 80.40           & 78.6                               & 35.8                                & 40.8                              \\ \midrule
    Ours$\ddagger$                                                  & \multicolumn{1}{c|}{\textbf{66.90}}            & \multicolumn{1}{c|}{89.00}        & 81.60           & \textbf{79.2}                               & \textbf{36.9}                                & \textbf{41.4}                              \\ \bottomrule
    \end{tabular}
\end{table}

\textbf{VEDAI Dataset.}
The experimental comparisons on the VEDAI dataset are given in Tab. \ref{tab_vedai}. Although we do not apply any tricks for small object detection, our method still outperforms the baseline method by 1.96$\%$ over mAP, and achieves competitive results with a mAP of 76.62$\%$ among the existing methods. However, under the more strict evaluation metrics mAP, our method is 0.28$\%$, 1.37$\%$ lower comparing to the Input Fusion and Mid Fusion method respectively.

\begin{table}[]
	\centering
	\caption{Comparison on the VEDAI Dataset.}
	\label{tab_vedai}
	\renewcommand\arraystretch{1}
	\setlength{\tabcolsep}{3pt}
	\scriptsize
	\begin{tabular}{l|c|c}
		\toprule
		\multicolumn{1}{c|}{\textbf{Method}} &  \textbf{mAP50(\%)} & \textbf{mAP(\%)} \\ \midrule

		Input Fusion \cite{qingyun2022cross}$\ddagger$                                                         & 74.40              & 45.65            \\
		Mid Fusion \cite{qingyun2022cross}$\ddagger$                                                            & 74.80              & \textbf{46.30}            \\
		SuperYOLO \cite{zhang2023superyolo}$\ddagger$                                                             & 73.61              & -                \\
		Ours(baseline)$\ddagger$                                                         & 74.66              & 44.09            \\
		Ours$\ddagger$                                                                   & \textbf{76.62}     & 44.93            \\ \bottomrule
	\end{tabular}
\end{table}

\subsection{Qualitative Analysis}
\begin{figure*}[!h]
    \centering
    \includegraphics[width=0.8\linewidth]{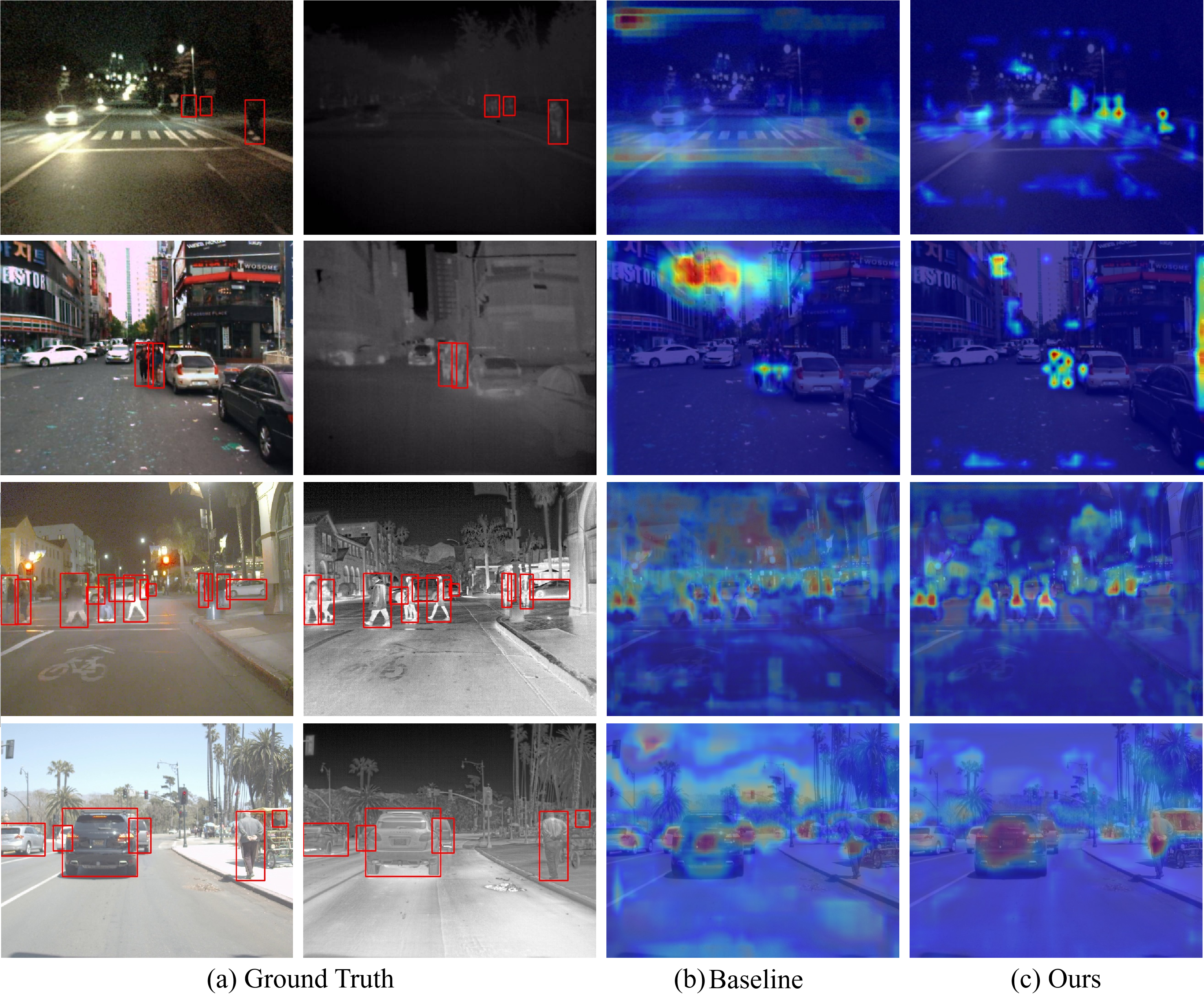}
    \caption{Visualization results of attention maps on KAIST and FLIR dataset. From left to right column: ground truth in RGB and thermal images, heatmaps of NIN fusion \cite{limultispectral} method (Baseline), and our proposed method.}
    \label{fig:Fig_visualkaist_FLIR}
\end{figure*}

Fig. \ref{fig:Fig_visualkaist_FLIR} illustrate sample visualization results of attention maps during daytime and nighttime on the KAIST and FLIR datasets. As shown in Fig. \ref{fig:Fig_visualkaist_FLIR}(a), it's difficult to detect the pedestrians under poor illumination conditions in the RGB images with naked eyes, however our method can still identify and locate the objects by aggregating RGB and thermal images. Besides, the complex urban traffic scenes mixed with pedestrians and vehicles bring great challenges, whereas our method is able to distinguish between different categories of objects with the assistance of auxiliary modality. 
Fig. \ref{fig:Fig_visualkaist_FLIR}(b) illustrate that the baseline method shows interest in different areas of the input images, resulting in more false positives. 
However, our method can exploit the global spatial location information and the correlation between different objects to capture highly-discriminative features, as shown in Fig. \ref{fig:Fig_visualkaist_FLIR}(c). 

\subsection{Limitations}
\label{appendix:C}
\begin{figure*}[h]
	\centering
	\includegraphics[width=1\linewidth]{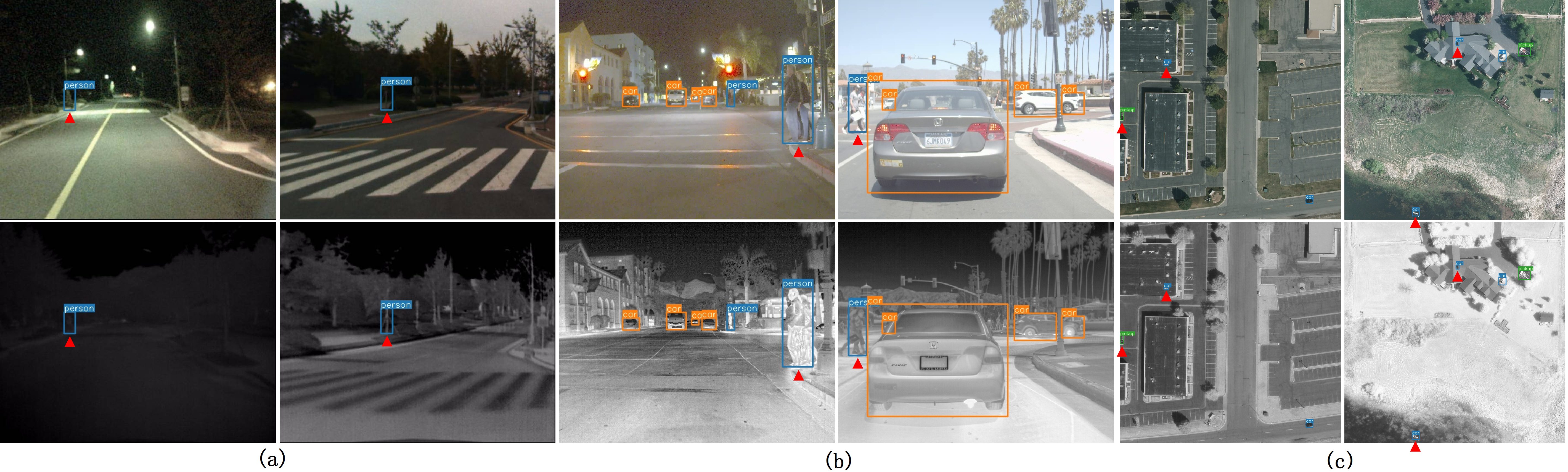}
	\caption{Failure cases on the KAIST, FLIR and VEDAI datasets. From  left to right columns are failure cases on the KAIST dataset (a), FLIR dataset (b) and VEDAI dataset (c). The red triangles indicate the false positives or false negatives in the images. Zoom in for more details.}
	\label{fig:Fig_failure}
\end{figure*}

In this section, we have provided some the failure cases and analyzed the limitations of our proposed method. Fig. \ref{fig:Fig_failure} (a) illustrates that our model misidentifies the traffic signs or trees as person in some scenarios. In our opinion, the main reasons for the false positives are the visual appearance similarity to the traffic signs or trees, and low image quality of the KAIST dataset. In Fig. \ref{fig:Fig_failure} (b), the occlusion between two overlapping pedestrians can also leads to false negatives on the FLIR dataset. Furthermore, Fig. \ref{fig:Fig_failure} (c) shows that our model may misidentify some devices mounted on rooftops as cars on the VEDAI dataset because they have similar shapes and colors when viewed from a bird's-eye perspective.

\section{Conclusions}
In this paper, we proposed a novel cross-modal feature fusion framework for multispectral object detection, which addresses the issue that exsiting methods mainly focus on local feature correlations between different modalities. More particularly, the cross-modal feature enhancement module is proposed to enhance the feature representation of mono-modality by leveraging the global information from complementary modality. Furthermore, we introduce iterative learning strategy to refine the complementary information, which improves the model performance without adding extra parameters. In terms of detection accuracy and running speed, our proposed method outperforms the other state-of-the-art methods on the KAIST, FLIR and VEDAI datasets. In the future, we intent to further explore the efficient and lightweight cross-modal feature fusion framework, and extent our approach to other multimodal tasks.

\section*{Acknowledgement}
This work was supported in part by NSF of China under Grant No. 61903164, 62276061 and in part by NSF of Jiangsu Province in China under Grants BK20191427. 

\section*{References}
\bibliography{mybibfile}


\end{document}